\documentclass{article}

\usepackage{PRIMEarxiv}

\usepackage[utf8]{inputenc} 
\usepackage[T1]{fontenc}    
\usepackage{hyperref}       
\usepackage{url}            
\usepackage{booktabs}       
\usepackage{amsfonts}       
\usepackage{nicefrac}       
\usepackage{xcolor}
\usepackage{microtype}      
\usepackage{lipsum}
\usepackage{amsmath}
\usepackage{fancyhdr}       
\usepackage{graphicx}       
\graphicspath{{media/}}     

\title{Survival Analysis Revisited: Understanding and Unifying Poisson, Exponential, and Cox Models in Fall Risk Analysis

}

\author{
  Tianhua Chen \\
  School of Computing and Engineering\\
  University of Huddersfield, UK \\
  \texttt{T.Chen@hud.ac.uk} \\
}

\begin{document}
\maketitle

\begin{abstract}

This paper explores foundational and applied aspects of survival analysis, using fall risk assessment as a case study. It revisits key time-related probability distributions and statistical methods, including logistic regression, Poisson regression, Exponential regression, and the Cox Proportional Hazards model, offering a unified perspective on their relationships within the survival analysis framework. A contribution of this work is the step-by-step derivation and clarification of the relationships among these models, particularly demonstrating that Poisson regression in the survival context is a specific case of the Cox model. These insights address gaps in understanding and reinforce the simplicity and interpretability of survival models. The paper also emphasizes the practical utility of survival analysis by connecting theoretical insights with real-world applications. In the context of fall detection, it demonstrates how these models can simultaneously predict fall risk, analyze contributing factors, and estimate time-to-event outcomes within a single streamlined framework. In contrast, advanced deep learning methods often require complex post-hoc interpretation and separate training for different tasks particularly when working with structured numerical data.
This highlights the enduring relevance of classical statistical frameworks and makes survival models especially valuable in healthcare settings, where explainability and robustness are critical. By unifying foundational concepts and offering a cohesive perspective on time-to-event analysis, this work serves as an accessible resource for understanding survival models and applying them effectively to diverse analytical challenges.
\end{abstract}

\keywords{ Time-To-Event Modelling \and Event Prediction \and Survival Analysis \and Risk Analysis \and Fall Risk Assessment \and Fall Prediction \and Logistic Regression \and Exponential Regression \and Poisson Regression \and Generalised Linear Models (GLMs) \and Cox Proportional Hazards Model \and Applied Statistics in Healthcare}

\section{Introduction}
Time-to-event analysis, often referred to as survival analysis \cite{hosmer2008applied,tolles2016time}, plays a crucial role in healthcare research, particularly in addressing the challenges posed by an aging population. As life expectancies increase, the prevalence of age-related health issues such as falls, chronic illnesses, cardiovascular conditions, and neurodegenerative diseases continues to rise. These conditions demand accurate modeling and prediction of event timing to design effective preventive strategies and improve patient outcomes. Survival analysis provides a robust framework for understanding and forecasting events like disease progression, survival times after diagnosis, and recovery durations.
For example, falls among the elderly population are often precursors to more severe conditions \cite
{clegg2013frailty,kim2024frailty}. Leveraging survival analysis to understand fall likelihood, timing, and associated risk factors enables early detection and intervention \cite{robinovitch2013video,bergen2021understanding,traverso2024investigating}, significantly enhancing patient care, especially for vulnerable elderly individuals.

The growing popularity of survival analysis in healthcare stems from its ability to model time-to-event outcomes in diverse real-world scenarios. At the same time, advances in deep learning , such as Recurrent Neural Networks (RNNs) and Variants (e.g., Long Short-Term Memory Networks (LSTMs), Gated recurrent units (GRUs)) \cite{bengio2016deep}, have introduced powerful data-driven tools for predictive modeling. However, despite their potential, these advanced methods often require substantial training data and computational resources and lack the interpretability and simplicity offered by traditional statistical approaches \cite{swamy2023future,marcinkevivcs2023interpretable}. In contexts like healthcare, where explainability and robustness are paramount, traditional methods retain critical advantages.
Another key advantage of statistical models lies in their ability to address multiple questions within a single framework. For example, a survival model can simultaneously predict fall risks at various time points, analyse the impact of covariates, and estimate the expected time until the next event. In contrast, deep learning approaches often require additional post-hoc methods for result interpretation, a process that can become even more complex when dealing exclusively with structured numerical data \cite{zhao2023interpretation}. Moreover, such models often require separate training for different tasks—for instance, one model to predict risk at specific time intervals and another to estimate the time to the next event, particularly when working with structured numerical clinical data, where the lack of inherent context or sequential relationships necessitates more tailored model architectures and training objectives \cite{borisov2022deep}, further increasing complexity and computational demands.

This paper is thus motivated by the need to revisit and unify foundational statistical methods in survival analysis, ensuring their continued relevance in the evolving analytical landscape. By providing a review and integration of widely used survival analysis techniques, the paper aims to clarify the connections between these models, address gaps in understanding their relationships, and offer a cohesive perspective for time-to-event analysis.

The discussion begins with logistic regression, a foundational method for binary classification. While effective for predicting outcomes at fixed time intervals, its limitations in modeling time-to-event data are evident. To address these shortcomings, the paper introduces Poisson and Exponential distributions, which incorporate time into predictive frameworks. The Generalized Linear Models are then explored as a bridge to enable the integration of covariates that are then adapted in survival analysis framework. Finally, the paper presents a comparative analysis of the Cox Proportional Hazards model, highlighting its strengths and its relationship with Poisson regression in the context of survival analysis.

In addition to theoretical insights, this paper emphasizes the practical applications of survival analysis methods. Using fall detection as a case study, it applies these models to achieve three main objectives: predicting the risk of an event (e.g., a fall) at specific intervals such as 3, 6, or 12 months, interpreting the influence of covariates on event risk, and estimating the expected time until the next event for an individual. These applications highlight the value of survival models in healthcare, providing actionable insights that support clinical decision-making and resource allocation.
\section{Predicting the Probability of an Event at a Fixed, Preselected Time Interval}

One of the fundamental questions in time-to-event prediction is estimating the probability of an event occurring within a specific, fixed time interval. For example, in the context of fall prediction, healthcare providers may ask: \textit{What is the probability of an elderly patient falling at Month 6 after a baseline assessment?} This is a natural and practical question that aligns with preventative care and resource allocation strategies.

Given a dataset of \( n \) observations, let \( X \in \mathbb{R}^p \) represent the vector of \( p \) predictor variables for an individual, and let \( Y \in \{0, 1\} \) denote the binary outcome indicating whether the event of interest occurs within the fixed time interval. The goal is to model the conditional probability:
\[
P(Y = 1 \mid X) = f(X),
\]
where \( f(X) \) maps the predictors to a probability value in the range \([0, 1]\). 

Logistic regression \cite{nick2007logistic,george2014survival} is a common approach to estimate \( f(X) \), which models the log-odds of the outcome as a linear function of the predictors:
\[
\log\left(\frac{P(Y = 1 \mid X)}{1 - P(Y = 1 \mid X)}\right) = \beta_0 + \beta_1 X_1 + \beta_2 X_2 + \dots + \beta_p X_p.
\]

The $\text{Odds} = \frac{P}{1-P}$ or the logarithm of the odds $\text{Log-odds} = \log\left(\frac{P}{1-P}\right)$, is modelled linearly in logistic regression, where each coefficient \( \beta_k \) is simple to interpret that represents the change in the log-odds for a one-unit increase in the factor \( X_k \), holding other predictors constant. The corresponding change in odds is: $\exp(\beta_k)$,
indicating the multiplicative change in the odds for a one-unit increase in \( X_k \). 

This can be rearranged to express the probability of the event as:
\[
P(Y = 1 \mid X) = \frac{1}{1 + \exp\left(-(\beta_0 + \beta_1 X_1 + \beta_2 X_2 + \dots + \beta_p X_p)\right)}.
\]
Logistic regression is computationally efficient and straightforward to implement, offering clear insights into the relationship between predictors and the likelihood of a fall. The model's coefficients indicate both the direction and magnitude of each predictor's influence on the outcome.

Several variations of logistic regression exist that are designed to handle issues like overfitting and variable selection, including Lasso Regression (L1 Regularization), Ridge Regression (L2 Regularization), and Elastic Net (combining L1 and L2 penalties) \cite{nick2007logistic,vsinkovec2021tune}. However, these methods share a common limitation when applied to time-to-event contexts, such as fall prediction:

\begin{enumerate}
    \item \textbf{Fixed Time Intervals:} Logistic regression requires a preselected time interval for modeling (e.g., predicting the probability of a fall at \( t = 6 \) months). While softmax regression can extend the output to more than two categories to handle multiple fixed intervals, these categories still need to be preselected and fixed in advance. This approach remains limited in its ability to address dynamic or continuous time intervals.
   
    \item \textbf{Lack of Temporal Dynamics:} The model does not explicitly incorporate time, failing to account for how risks evolve over different time points.

    \item \textbf{No Consideration of Event Timing:} Logistic regression treats the outcome as binary, ignoring when the event occurs within the interval. This is a significant drawback in scenarios like fall prediction, where the timing of events is critical for planning interventions.
\end{enumerate}

Logistic regression serves as a solid foundation for modeling the probability of an event within a fixed time interval. However, its inability to model temporal dynamics or varying risks over time underscores the need for alternative approaches. The following sections address these limitations by explicitly incorporating the time element and explore survival analysis framework into the modeling process.

\section{Modelling Time Element with Probability Distributions}

Essentially a binary outcome without accounting for when the event occurs, the primary drawback of logistic regression lies in its inability to incorporate time in the model. To address this limitation in time-related events, it is natural to consider probability distributions that explicitly model time. Two commonly used distributions in this context are the Exponential distribution and the Poisson distribution \cite{ross2014introduction}.

\textbf{Poisson distribution:} The Poisson distribution models the number $k$ of events (\( N_t \)) occurring in a time interval \( t \), given a constant event rate \( \lambda \):
    \[
    P(N_t = k) = \frac{(t \lambda)^k \exp(-\lambda t)}{k!}, \quad k = 0, 1, 2, \dots
    \]
\textbf{Exponential distribution:} The Exponential distribution models the time (\( T \)) between consecutive events, given a constant hazard rate \( \lambda \). Its probability density function (PDF) and cumulative distribution function (CDF) are:
    \[
    f(T) = \lambda \exp(-\lambda T), \quad T \geq 0,
    \]
    \[
    F(T) = P(T \leq t) = 1 - \exp(-\lambda T).
    \]

It is natural to align the choice of distribution with specific research questions related to fall prediction. Below are two possible key questions of interest:

\begin{itemize}
    \item \textbf{Q1:} What is the probability that a fall occurs within \( t \) months?
    \item \textbf{Q2:} What is the probability that no falls occur until at least \( t \) months later?
\end{itemize}

These questions highlight different perspectives on modeling time-to-event data, with an analysis as follows.

\section*{Q1: What is the probability of a fall occurring within \( t \) months?}

To address this question, we consider the probability of one or more falls occurring within a fixed time interval \( t \). 

\subsection*{Case 1-1: Exactly One Fall}
The Poisson distribution models the number of events occurring in a fixed time interval \( t \), with rate parameter \( \lambda \). The probability of observing exactly one fall within \( t \) units of time (such as months) is:

\[
P(N_t = 1) = \frac{(t \lambda)^1 \exp(-\lambda t)}{1!} = t \lambda \exp(-\lambda t).
\]

This assumes that falls follow a Poisson process, where events occur independently, and the rate \( \lambda \) is constant.

\subsection*{Case 1-2: One or More Falls}
The probability of one or more falls within \( t \) months is the complement of the probability of no falls (\( N_t = 0 \)):

\[
P(N_t \geq 1) = 1 - P(N_t = 0).
\]

From the Poisson distribution:
\[
P(N_t = 0) = \frac{(t \lambda)^0 \exp(-\lambda t)}{0!} = \exp(-\lambda t).
\]

Thus, the probability of one or more falls is:
\[
P(N_t \geq 1) = 1 - \exp(-\lambda t).
\]

\subsection*{From the Perspective of Exponential Distribution}
The exponential distribution, which models the time between events, can also answer this question. The cumulative distribution function (CDF) of the exponential distribution is:
\[
P(T \leq t) = 1 - \exp(-\lambda t),
\]
where \( T \) is the time to the first event (e.g., a fall). This result is consistent with the Poisson-derived probability of one or more events occurring within \( t \) months:
\[
P(N_t \geq 1) = P(T \leq t) = 1 - \exp(-\lambda t).
\]

The equivalence \( P(T \leq t) = P(N_t \geq 1) \) arises from the relationship between the exponential and Poisson distributions in a Poisson process. While mathematically identical, their interpretations differ. \( P(T \leq t) \) models the probability of the first event occurring within time \( t \), focusing exclusively on the timing of that first event, regardless
of what happens afterward, leaving subsequent events unmodeled. 
This distinction becomes explicit in the Poisson distribution, where probabilities for specific event counts, such as \( P(N_t = 1) \) or \( P(N_t = 2) \), are directly modeled. However, when using the exponential distribution, the focus is inherently on the first event or the interval between consecutive events, aligning it with time-focused questions rather than event-count-focused scenarios. For applications requiring explicit event counts, such as determining the probability of exactly one event occurring within \( t \), the Poisson distribution provides the necessary framework. The equivalence \( 1 - e^{-\lambda t} \) thus serves different purposes depending on the context: it can represent the probability of the first event occurring or the probability of at least one event, but caution is needed to ensure correct interpretation based on the specific problem at hand.

\section*{Q2: What is the probability that no falls occur until \( t \) months later?}

The second question shifts the focus to the survival probability \( S(t) \), a key concept in survival or risk analysis \cite{tolles2016time}, which quantifies the likelihood of not experiencing an event, such as a fall, by time \( t \).

\subsection*{Case 2-1: Based on the Poisson Distribution}
The probability of no falls occurring within \( t \) months is equivalent to \( P(N_t = 0) \), calculated earlier:
\[
S(t) = P(T \geq t) = P(N_t = 0) = \exp(-\lambda t).
\]

\subsection*{Case 2-2: Based on the Exponential Distribution}
The exponential distribution, particularly the cumulative form of the exponential distributions, answers 
\[
P(T \leq t) = 1-P(T \geq t) = 1-\exp(-\lambda t) = 1-S(t)
\]
Thus:
\[
S(t) =  P(T \geq t) = \exp(-\lambda t)
\]
The equivalence between the Poisson-derived survival function and the exponential distribution-derived survival function underscores their inherent relationship. Both describe the same survival probability under the assumption of a constant hazard rate \( \lambda \). 

The second question—what is the probability that no falls occur until \( t \) months later?—is particularly relevant in the context of fall prediction. This type of modeling, referred to as survival analysis or risk analysis, shifts the focus from the occurrence of an event within a fixed interval to understanding the likelihood of avoiding the event entirely up to a specific time \cite{hosmer2008applied,tolles2016time}. Predicting the survival probability \( S(t) \) enables a dynamic understanding of risk, particularly for time-sensitive outcomes.
Understanding the likelihood of maintaining a safe state (e.g., no falls) provides actionable insights into intervention timing and efficacy.
Following sections will delve into survival analysis methods that enable robust predictions of \( S(t) \) while incorporating covariates to refine individual risk assessments.

\section{Poisson Regression in Survival Analysis Framework}
Now that we can model the time of interest using probability distributions such as the Poisson and Exponential distributions, which inherently account for time; however, they do not directly incorporate the influence of covariates. To address this limitation, we utilise Generalized Linear Models (GLMs) \cite{matloff2017statistical,dobson2018introduction}, which enable the modeling of the linear relationship between the risk of time-to-event and associated covariates. 

\subsection*{The Exponential Family and Its Role in GLMs}

A prerequisite to understanding GLMs is the exponential family of distributions, a versatile class of probability distributions expressed in the form:
\[
f(y; \eta) = b(y) \exp\left(\eta^\top T(y) - a(\eta)\right),
\]
where:
\begin{itemize}
    \item \( \eta \): the natural (canonical) parameter, often derived by rewriting the original parameters of the distribution,
    \item \( T(y) \): the sufficient statistic, summarizing all relevant information about \( y \) (commonly \( T(y) = y \)),
    \item \( a(\eta) \): the log-partition function, ensuring normalization,
    \item \( b(y) \): the base measure, independent of \( \eta \).
\end{itemize}

Rewriting a distribution into this form typically involves taking the logarithm of the probability density function (PDF) or probability mass function (PMF) and identifying the components corresponding to \( \eta \), \( T(y) \), and \( a(\eta) \). The following are examples of common distributions in exponential family form, which provide the foundation for deriving flexible linear models, where the response variable may follow alternative distributions depending on its nature.
\begin{itemize}
    \item \textbf{Bernoulli: $f(y; p) = p^y (1-p)^{1-y}, \quad y \in \{0, 1\}, \, 0 < p < 1$}
     \[
    f(y; \eta) = \exp\left(y \eta - \ln(1 + e^\eta)\right), \text{where } \eta= \ln\left(\frac{p}{1-p}\right), a(\eta) =  \ln(1 + e^\eta),  b(y) = 1.
    \]
     \item \textbf{Exponential: $f(y; \lambda) = \lambda e^{-\lambda y}, \quad y \geq 0, \, \lambda > 0$}
     \[ f(y; \eta) = \exp\left(\eta y - \ln(-\eta)\right), \text{where } \eta= -\lambda, a(\eta) =  \ln(-\eta),  b(y) = 1.
    \]
    \item \textbf{Poisson: $f(y; \lambda) = \frac{\lambda^y e^{-\lambda}}{y!}, \quad y \in \{0, 1, 2, \dots\}, \, \lambda > 0.$}
    \[
    f(y; \eta) = \exp\left(\eta y - e^\eta - \ln(y!)\right), \text{where } \eta = \ln(\lambda), a(\eta) =  e^\eta,  b(y) = \frac{1}{y!}
    \]
\end{itemize}

\subsection*{GLMs: Associating Covariates with Risks}

GLMs extend linear regression to allow the response variable \( Y \) to follow distributions from the exponential family. 
To derive GLMs, three core principles are followed:

\paragraph{1. Response Variable Belongs to the Exponential Family:} The response variable \( Y \) must follow an exponential family distribution:$
f(y; \eta) = b(y) \exp\left(\eta^\top T(y) - a(\eta)\right).$

\paragraph{2. Model Predicts the Expected Value of \( Y \):} GLMs predict the expected value of \( Y \) given predictors \( X \):$h_\beta(X) = \mathbb{E}[Y | X].$
The relationship between the linear predictor \( \eta \) and the expected value \( h_\beta(X) \) is established through a link function, which transforms \( \eta \) to match the domain of \( Y \). 

\paragraph{3. Linear Relationship in Predictors:}
The linear predictor \( \eta \) is modeled as:$
\eta = \beta^\top X.$

The following examples illustrate GLMs applied to different distributions:

\paragraph{Logistic Regression:}
For binary outcomes modeled by a Bernoulli distribution, which essentially results in the logistic regression as applied in Section 2:
\[
h_\beta(X) = \frac{1}{1 + \exp(-\beta^\top X)}.
\]

\paragraph{Exponential Regression:}
For time-to-event outcomes modeled by an exponential distribution:
\[
h_\beta(X) = \frac{1}{-\beta^\top X}.
\]
where \( \beta^\top X < 0 \) is required to ensure \( h_\beta(X) > 0 \), as time-to-event outcomes must always be positive. This constraint introduces additional challenges during model training, as it requires careful handling to enforce the negativity of \( \beta^\top X \) consistently, and is thus not preferred in practice.

\paragraph{Poisson Regression:}
Based on the three conditions for deriving Poisson regression through GLMs, \( h_\beta(X) \) represents the expected value \( E[Y|X] \), which corresponds to the mean \( \lambda \) of the Poisson-distributed response variable \( Y \). By rewriting the Poisson distribution in its exponential family form, it follows that:
\[
\lambda = \exp(\eta),
\]
where \( \eta = \beta^\top X \). Consequently, the Poisson regression model can be expressed as:
\[
h_\beta(X) = \exp(\beta^\top X).
\]

\section*{Survival Analysis Using Poisson Regression}

The survival function \( S(t) \) derived in Section 3 is given by:  
\[
S(t) = P(T \geq t) = \exp(-\lambda t),
\]
where \( \lambda \) represents the event rate, consistent across both the exponential and Poisson distributions.

By leveraging Poisson regression derived through GLMs, the event rate \( \lambda \) can be expressed as a function of covariates:
\[
\lambda = \exp(\eta) = \exp(\beta^\top X).
\]
Substituting this expression for \( \lambda \) into the survival function results in:
\[
S(t) = \exp(-t \exp(\beta^\top X)).
\]
This formulation represents a survival model, linking the survival probability to covariates through GLMs.

\section{Cox Proportional Hazard Model}
The previous section discussed how to utilize common probability distributions and Generalized Linear Models (GLMs) to derive Poisson regression for time-to-event analysis within the framework of survival analysis. Before a comparative analysis, this section introduces a widely used approach for analyzing time-to-event data: the Cox Proportional Hazards Model (Cox PH Model) \cite{cox1972regression}. It models key concepts, including the hazard rate, cumulative hazard function, and survival function, while seamlessly incorporating covariates to account for individual differences.

\subsection*{Basic Concepts}
The hazard rate represents the instantaneous risk $h(t)$ of the event occurring at time \( t \), given survival up to that time:
\[
h(t) = \lim_{\Delta t \to 0} \frac{P(t \leq T < t + \Delta t \, | \, T \geq t)}{\Delta t}.
\]
It's worth noting that the hazard rate is not a probability but a \textit{rate} (e.g., events per unit time).

The cumulative hazard function \( H(t) \) measures the total accumulated risk of experiencing the event up to time \( t \).
\[
H(t) = \int_0^t h(u) \, du.
\]

The survival function \( S(t) \) gives the probability of surviving (not experiencing the event) beyond time \( t \):
\[
S(t) = P(T \geq t).
\]

From the definition of the hazard function, the probability of surviving a small time interval \( [t, t + \Delta t) \), given survival up to \( t \), is approximately:
\[
P(T \geq t + \Delta t \, | \, T \geq t) = 1 - h(t) \Delta t.
\]
Taking the product over all infinitesimal intervals up to \( t \), the survival probability becomes:
\[
S(t) = \prod_{u=0}^t \big(1 - h(u) \Delta u\big).
\]
Using the approximation \( \ln(1 - x) \approx -x \) for small \( x \), and taking the logarithm:
\[
\ln(S(t)) \approx \sum_{u=0}^t -h(u) \Delta u = -\int_0^t h(u) \, du.
\]
Exponentiating both sides:
\[
S(t) = \exp\big(-H(t)\big).
\]

While \( H(t) \) captures the total accumulated risk of the event up to time \( t \), and \( \exp(-H(t)) \) translates this cumulative risk into the probability of not experiencing the event. Since \( H(t) \geq 0 \), \( S(t) = \exp(-H(t)) \) ensures \( 0 \leq S(t) \leq 1 \), satisfying the requirements of a probability. Survival over \( t \) is the product of surviving each infinitesimal time interval, and \( \exp(-H(t)) \) reflects this multiplicative accumulation.

\subsection*{Incorporating Covariates: Proportional Hazards Assumption}

The Cox PH model relates the hazard rate to a set of covariates as:
\[
h(t | X) = h_0(t) \cdot \exp(\beta^\top X),
\]
where \( h_0(t) \) is the baseline hazard function and \( \exp(\beta^\top X) \) is the relative risk associated with the covariates \( X \). This decomposition separates the hazard function into two distinct components:
the baseline hazard \( h_0(t) \), which depends solely on time \( t \) and captures the risk for an individual when all covariates \( X = 0 \); and the relative risk \( \exp(\beta^\top X) \), which adjusts the baseline hazard based on the covariates and remains constant over time.

The use of \( \exp(\beta^\top X) \) appears identical to that of Poisson that is derived from GLMs, but in the context of Cox model, the use of \( \exp(\beta^\top X) \) is more of a design choice to ensure non-negative hazards, interpretable coefficients, and realistic multiplicative effects.
\begin{enumerate}
    \item 
    \textbf{Ensuring Non-Negativity of the Hazard:} The hazard rate \( h(t | X) \) represents a risk or rate and must always be non-negative. The exponential function \( \exp(\cdot) \) naturally satisfies this requirement, as \( \exp(z) > 0 \) for all \( z \). This is a crucial advantage over direct linear formulations such as:
\[
h(t | X) = h_0(t) + \beta^\top X,
\]
which can result in invalid negative hazard rates when \( \beta^\top X \) dominates \( h_0(t) \). 

\item \textbf{Logarithmic Interpretation}
The exponential function introduces a log-linear relationship between covariates and the hazard, making the model interpretable:
\[
\log(h(t | X)) = \log(h_0(t)) + \beta^\top X.
\]
where each coefficient \( \beta_k \) represents the log-hazard ratio associated with a one-unit increase in the \( k \)-th covariate, holding all other covariates constant.
For example:
If \( \beta_k = 0.5 \), then \( \exp(0.5) \approx 1.65 \), meaning a 65\% increase in the hazard for a one-unit increase in the covariate.

\item \textbf{Multiplicative Nature of the Hazard}
In the Cox model, covariates affect the hazard rate multiplicatively:
\[
h(t | X) = h_0(t) \cdot \exp(\beta^\top X) = h_0(t) \cdot  \exp(\beta_1 X_1) \cdot\exp(\beta_2 X_2) \cdots \cdot\exp(\beta_p X_p).
\]
This assumption aligns with many real-world scenarios where risks combine proportionally rather than additively should it formulate via direct linear model. For example, the risk of falling may increase proportionally with factors like age or mobility impairment, relative to a baseline hazard.
\end{enumerate}
It is worth noting that \( h_0(t) \) does not assume a specific form for the baseline hazard, making it a semi-parametric model. This is complemented by the use of \( \exp(\beta^\top X) \), which facilitates the estimation of \( \beta \) through the partial likelihood \cite{sinha2003bayesian}, a computationally efficient approach that avoids the need to specify the baseline hazard. This is made possible due to that the hazard ratio between two individuals remains constant over time:

\[
\frac{h(t \mid X_1)}{h(t \mid X_2)} = \exp\left(\beta^\top (X_1 - X_2)\right).
\]

This ratio depends on \( \beta \) and the covariates, not on \( t \), and remain constant regardless of \( t \). Then the cumulative hazard function that incorporates the covariates is defined as:
\[
H(t | X) = H_0(t) \exp(\beta^\top X),
\]
where:
\[
H_0(t) = \int_0^t h_0(u) \, du
\]
is the baseline cumulative hazard. Given \( H(t | X) \), the survival function is updated as:
\[
S(t | X) = \exp\big(-H(t | X)\big) = \exp\big(-H_0(t) \exp(\beta^\top X)\big).
\]

\section{Comparison of Poisson Regression and Cox PH Model}

Both Poisson regression applied in a survival analysis context and the Cox model aim to model time-to-event data. While their formulas are structurally similar, they differ in how the hazard rate and survival function are defined. This section reviews and compares these approaches.

\subsection*{1. Poisson Regression in a Survival Context}

\textbf{Hazard Rate:}  
In Poisson regression, the rate of events (\( \lambda \)) is modeled as:
\[
\lambda = \exp(\beta^\top X),
\]
where \( \beta^\top X \) is the log-linear combination of covariates.

\textbf{Survival Function:}  
For a constant event rate (\( \lambda = \exp(\beta^\top X) \)), the survival function is:
\[
S(t | X) = \exp(-t \lambda) = \exp\left(-t \exp(\beta^\top X)\right).
\]
This survival function is derived from the Poisson process based assumption for inter-event times.

\subsection*{2. Cox Proportional Hazards Model with Constant Baseline Hazard}

\textbf{Hazard Rate:}  
The Cox PH model specifies the hazard function as:
\[
h(u | X) = h_0(u) \exp(\beta^\top X).
\]
When the baseline hazard \( h_0(u) \) is constant (\( h_0(u) = \lambda_0 \)), the hazard becomes:
\[
h(u | X) = \lambda_0 \exp(\beta^\top X).
\]
Here, \( \lambda_0 \) acts as the constant baseline rate, while \( \exp(\beta^\top X) \) adjusts the hazard based on covariates.

\textbf{Survival Function:}  
The survival function in the Cox model is:
\[
S(t | X) = \exp\left(-\int_0^t h(u | X) \, du\right) = \exp\left(-\int_0^t h_0(u) exp(\beta^T X) \, du\right).
\]
For a constant hazard (\( h(u | X) = \lambda_0 \exp(\beta^\top X) \)), this simplifies to:
\[
S(t | X) = \exp\left(-t \lambda_0 \exp(\beta^\top X)\right).
\]

\subsection*{Comparison of Formulas}

\textbf{Hazard Rate:}
\begin{itemize}
    \item \textbf{Poisson Regression:} Directly models the rate parameter \( \lambda = \exp(\beta^\top X) \).
    \item \textbf{Cox PH Model:} Separates the rate into the effect of covariates \( \exp(\beta^\top X) \) and a baseline hazard $h_0(t)$, which is a function to time $t$, but can be considered constant in a simplified assumption:
   $\lambda = \lambda_0 \exp(\beta^\top X).$
\end{itemize}

\textbf{Survival Function:}
\begin{itemize}
    \item \textbf{Poisson Regression: $S(t | X) = \exp\left(-t \exp(\beta^\top X)\right).$}
    \item \textbf{Cox PH Model with constant baseline hazard assumption: $    S(t | X) = \exp\left(-t \lambda_0 \exp(\beta^\top X)\right).$}
\end{itemize}

As such, the main difference lies in the baseline hazard: In Poisson regression, the baseline hazard is implicitly absorbed into \( \exp(\beta^\top X) \); in the Cox model, the baseline hazard \( \lambda_0 \) is explicitly modeled, even when constant. If \( \lambda_0 = 1 \) in the Cox model, its formula for the survival function becomes identical to that of Poisson regression. In a nutshell, the Cox PH model is a more general approach, allowing for time-varying baseline hazards through $\int_0^t h_0(u) \, du$, whereas Poisson regression assumes a constant hazard over time. In this sense, Poisson regression can be seen as a special case of the Cox model under the assumption of a constant baseline hazard. 
It is important to note that time-varying baseline hazards differ from the time-invariant coefficients associated with covariates, which remain fixed over the period under consideration in both models.

\section{Learning and Applying Cox Proportional Hazard Model}
Given the broader applicability of the Cox Proportional Hazards model compared to Poisson regression in survival analysis, this section introduces the fundamentals of learning the Cox model and demonstrates its application in scenarios such as fall detection through adding three key applications: 1) Predicting the risk of an event at specific time intervals, 2) Analysing the influence of factors on event risk, and 3) Estimating the expected time until the next event.
\subsection{Learning a Survival Model}

To learn a survival model like the Cox PH model, the key is to estimate the relationships between covariates and survival time (via the hazard function), while simultaneously capturing the underlying baseline hazard and survival probabilities. Below are core computations involved:

\subsubsection*{Step 1: Estimate the Regression Coefficients (\( \beta \))}

The first task is to learn how the covariates \( X \) influence the hazard via the relative risk:
\[
h(t | X) = h_0(t) \exp(\beta^\top X).
\]

The coefficients \( \beta \) are estimated using  partial likelihood, which focuses on the ordering of event times rather than their exact values. Maximizing partial likelihood gives estimates for \( \beta \), capturing the effect of covariates on the hazard ratio.
\[
L(\beta) = \prod_{i=1}^n \frac{\exp(\beta^\top X_i)}{\sum_{j \in R(t_i)} \exp(\beta^\top X_j)},
\]
where: \( t_i \) is the time of the event for individual \( i \), and \( R(t_i) \) is the set of individuals still at risk just before \( t_i \). Using partial likelihood makes it unnecessary to compute the baseline hazard \( h_0(t) \), as it cancels out in the partial likelihood.

\subsubsection*{Step 2: Estimate the Baseline Cumulative Hazard (\( H_0(t) \))}

After \( \beta \) is estimated, the baseline cumulative hazard \( H_0(t) \), which accumulates the risk of events over time in the baseline population (i.e., individuals with \( X = 0 \)), is then computed by aggregating \( \Delta H_0(t_i) \) over all distinct event times \( t_1, t_2, \dots, t_k \), to get the cumulative hazard:
\[
H_0(t) = \sum_{t_i \leq t} \Delta H_0(t_i).
 \]
where $\Delta H_0(t_i)$ is defined below with \( d_i \) being the number of events at time \( t_i \), \( R(t_i) \) being the risk set at \( t_i \).
 \[
\Delta H_0(t_i) = \frac{d_i}{\sum_{j \in R(t_i)} \exp(\beta^\top X_j)},
\]

\subsubsection*{Step 3: Compute the Baseline Survival Probability (\( S_0(t) \))}

Once \( H_0(t) \) is computed, the baseline survival probability is derived as:
\[
S_0(t) = \exp(-H_0(t)).
\]
The \( S_0(t) \) gives the probability of surviving beyond time \( t \) for an individual with \( X = 0 \), and forms the foundation for calculating survival probabilities for individuals with covariates.

\subsubsection*{Step 4: Compute Survival Probabilities for Individuals}
To compute the survival probability for an individual with covariates \( X \), adjust the baseline survival probability using the covariate effect:
\[
S(t | X) = S_0(t)^{\exp(\beta^\top X)}  = \exp\big(-H_0(t) \exp(\beta^\top X)\big).
\]

\subsection{Applying the Survival Model}
Once a survival model, such as the Cox model or Poisson regression, is trained, it can be applied to various time-to-event analyses. This section highlights three key applications:
\begin{enumerate}
    \item Predicting the risk of an event (e.g., fall) for an individual at specific time intervals, such as 3, 6, and 12 months.
    \item Interpreting the influence of different factors on the risk of the event.
    \item Estimating the expected time until a specific individual experiences the next event.
\end{enumerate}

\subsection*{1. Predicting Survival Probabilities at Specific Time Points}

The survival probability at time \( t \), given a set of covariates \( X \), is defined as:
\[
S(t | X) = \exp\left(-H(t | X)\right),
\]
where \( H(t | X) \) is the cumulative hazard, computed as:
\[
H(t | X) = H_0(t) \cdot \exp(\beta^\top X).
\]
Here, \( H_0(t) \) represents the baseline cumulative hazard, and \( \beta^\top X \) is the linear predictor derived from the model coefficients \( \beta \) and covariates \( X \).

To predict survival probabilities:
\begin{enumerate}
    \item Compute the linear predictor:
    \[
    \beta^\top X = \sum_{i} \beta_i X_i.
    \]
    \item Evaluate the cumulative hazard:
    \[
    H(t | X) = H_0(t) \cdot \exp(\beta^\top X).
    \]
    \item Calculate the survival probability:
    \[
    S(t | X) = \exp\left(-H(t | X)\right).
    \]
\end{enumerate}

This method can be applied to estimate survival probabilities at various time points, such as \( t = 3 \), \( t = 6 \), or \( t = 12 \) months, for individuals with specific covariates.

\subsection*{2. Estimating Time to an Event}

To estimate the time \( t \) until an event occurs, use the survival probability \( S(t | X) \). The time at which \( S(t | X) \) reaches a specific threshold, such as 50\% that may represent the median survival time, can be computed as follows:

\[
S(t | X) = \exp\left(-H(t | X)\right),
\]
where the cumulative hazard \( H(t | X) \) is:
\[
H(t | X) = H_0(t) \cdot \exp(\beta^\top X).
\]

For \( S(t | X) = 0.5 \), the cumulative hazard is:
\[
H(t | X) = -\ln(0.5) \approx 0.693.
\]
Rearranging gives:
\[
H_0(t) = \frac{0.693}{\exp(\beta^\top X)}.
\]

Using the baseline cumulative hazard table \( H_0(t) \), the corresponding \( t \) can be identified. 
The table enables quick conversion of cumulative hazard values into survival times, providing a practical way to estimate time-to-event for individuals.
The \( H_0(t) \) can be derived as per Section 7.1, with an example below to demonstrate its usage for estimating time until next event.

\begin{center}
\begin{tabular}{|c|c|}
\hline
\textbf{Time (Months)} & \( H_0(t) \) \\ \hline
1 & 0.10 \\ \hline
2 & 0.25 \\ \hline
3 & 0.40 \\ \hline
4 & 0.60 \\ \hline
5 & 0.85 \\ \hline
6 & 1.10 \\ \hline
\end{tabular}
\end{center}

Assume a patient with \( \beta^\top X = 2.0 \), the cumulative hazard is scaled by \( \exp(2.0) = 7.39 \). Solving for \( H_0(t) \):
\[
H_0(t) = \frac{0.693}{7.39} \approx 0.094.
\]
From the table, \( H_0(t) \approx 0.094 \) corresponds to slightly before 1 month. Thus, the median survival time is approximately 1 month.

\subsection*{3. Interpreting Risks Associated with Covariates}

The coefficients \( \beta \) in the survival model represent the influence of each covariate \( X_i \) on the hazard function. Specifically, the hazard function is expressed as:
\[
h(t | X) = h_0(t) \cdot \exp(\beta^\top X).
\]
For a given covariate \( X_i \):
\begin{itemize}
    \item A positive \( \beta_i \) indicates that an increase in \( X_i \) raises the hazard, meaning the event is more likely to occur sooner, which corresponds to reduced survival.
    \item A negative \( \beta_i \) implies that an increase in \( X_i \) lowers the hazard, suggesting a protective effect that prolongs survival.
\end{itemize}

The magnitude of the effect is interpreted through the hazard ratio associated with a one-unit increase in \( X_i \):
\[
\text{Hazard Ratio} = \exp(\beta_i).
\]
\begin{itemize}
    \item If \( \exp(\beta_i) > 1 \), the risk increases with \( X_i \). For example, if \( \beta_i = 0.5 \), then \( \exp(\beta_i) \approx 1.65 \), meaning the hazard increases by 65\% for each unit increase in \( X_i \).
    \item If \( \exp(\beta_i) < 1 \), the risk decreases with \( X_i \). For example, if \( \beta_i = -0.5 \), then \( \exp(\beta_i) \approx 0.61 \), meaning the hazard decreases by 39\% for each unit increase in \( X_i \).
\end{itemize}

This interpretation allows for a clear understanding of how different factors influence the likelihood and timing of the event under study.

\section{Conclusion}
This paper underscores the critical role of survival analysis in tackling healthcare challenges, with a specific focus on fall detection and prevention. By revisiting foundational statistical methods and distributions, it offers a unified perspective on widely used survival analysis techniques, including logistic regression, Poisson and Exponential distributions, Generalized Linear Models (GLMs), and the Cox Proportional Hazards model.

A contribution of this work is the clarification of the relationships among these models, particularly demonstrating that Poisson regression in the survival context is a specific case of the Cox model. These insights bridge gaps in understanding and reinforces the simplicity, interpretability, and versatility of survival models. Unlike advanced deep learning methods—which often require complex post-hoc interpretation and separate training for different tasks especially when dealing with pure numerical data—survival models offer a streamlined framework capable of simultaneously predicting event risks at specific time intervals, interpreting covariate effects, and estimating time to the next event. These attributes are particularly valuable in healthcare, where explainability and robustness are paramount, making the applications are not limited to fall detection alone, but to broader domains such as disease progression, highlighting the versatility of survival analysis techniques.

By connecting theoretical insights with real-world applications, the paper emphasises the enduring relevance of classical statistical survivla frameworks in guiding intervention strategies and navigating the evolving landscape of analytical methods.
By detailing the derivation and application of survival models, the paper provides an accessible resource for understanding these techniques, making it a valuable entry point for researchers and practitioners alike. For further exploration of survival analysis and the Cox model, readers may refer to a series of tutorial papers on survival analysis \cite{clark2003survival,bradburn2003survival,bradburn2003survival2,clark2003survival3} and review papers on the Cox model \cite{kumar1994proportional,kalbfleisch2023fifty}.

\bibliographystyle{unsrt}  
\bibliography{references.bib} 
\end{document}